\documentclass[runningheads]{llncs}
\usepackage{graphicx}
\usepackage{tabularx}
\usepackage{hyperref}
\usepackage{color}

\usepackage{array}
\usepackage{multirow}
\usepackage{CJKutf8}
\usepackage{amsmath}
\usepackage{amsfonts}
\usepackage{amssymb}

\newcommand{\redb}[1]{\textcolor{red}{\textbf{#1}}}
\newcommand{\blueu}[1]{\textcolor{blue}{\underline{#1}}}

\newcolumntype{L}{>{\raggedright\arraybackslash}X}
\newcolumntype{R}{>{\raggedleft\arraybackslash}X}
\newcolumntype{C}{>{\centering\arraybackslash}X}

\DeclareMathOperator*{\argmax}{argmax}

\begin{document}
\title{CED: Catalog Extraction from Documents}
\titlerunning{CED: Catalog Extraction from Documents}
%
\author{
    Tong Zhu\inst{1} \and
    Guoliang Zhang\inst{1} \and
    Zechang Li\inst{2} \and
    Zijian Yu\inst{1} \\
    Junfei Ren\inst{1} \and
    Mengsong Wu\inst{1} \and
    Zhefeng Wang\inst{2} \\
    Baoxing Huai\inst{2} \and
    Pingfu Chao\inst{1} \and
    Wenliang Chen\inst{1}\thanks{Corresponding author}
}
%
\authorrunning{Zhu et al.}
%
\institute{
    Institute of Artificial Intelligence, School of Computer Science and Technology, Soochow University, China \and Huawei Cloud, China \\
    \email{\{tzhu7,glzhang,zjyu,jfrenjfren,mswumsw\}@stu.suda.edu.cn}\\
    \email{\{lizechang1,wangzhefeng,huaibaoxing\}@huawei.com} \\
    \email{\{pfchao,wlchen\}@suda.edu.cn}
}
%
\maketitle              

\begin{abstract}
Sentence-by-sentence information extraction from long documents is an exhausting and error-prone task.
As the indicator of document skeleton, catalogs naturally chunk documents into segments and provide informative cascade semantics, which can help to reduce the search space.
Despite their usefulness, catalogs are hard to be extracted without the assist from external knowledge.
For documents that adhere to a specific template, regular expressions are practical to extract catalogs.
However, handcrafted heuristics are not applicable when processing documents from different sources with diverse formats.
To address this problem, we build a large manually annotated corpus, which is the first dataset for the Catalog Extraction from Documents (CED) task.
Based on this corpus, we propose a transition-based framework for parsing documents into catalog trees.
The experimental results demonstrate that our proposed method outperforms baseline systems and shows a good ability to transfer.
We believe the CED task could fill the gap between raw text segments and information extraction tasks on extremely long documents.
Data and code are available at \url{https://github.com/Spico197/CatalogExtraction}


\end{abstract}

\section{Introduction}

\begin{figure}[t]
    \centering
    \includegraphics[width=\textwidth]{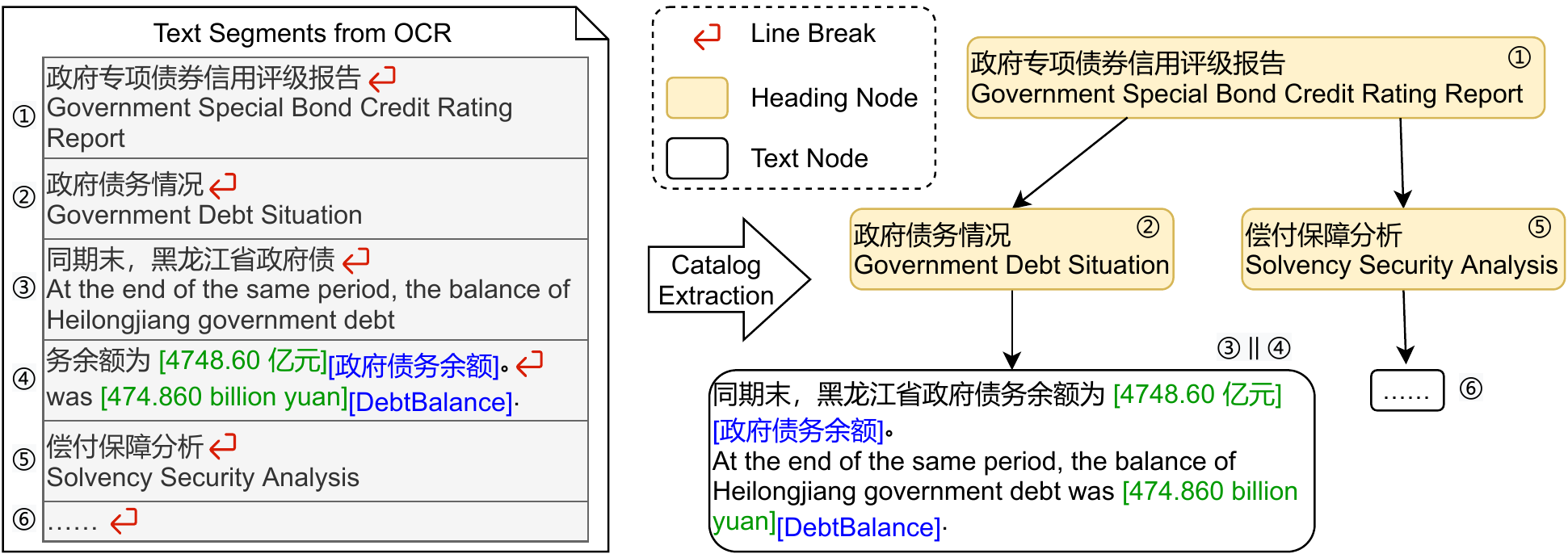}
    \caption{An example of catalog extraction. The text segments on the left are converted to a catalog tree on the right. The third and the fourth segments are concatenated after catalog extraction.}
    \label{fig:ced_example}
\end{figure}

Information in long documents is usually sparsely distributed \cite{yang-etal-2018-dcfee,zhu-et-al-ptpcg}, so a preprocessing step that distills the structure is necessary to help reduce the search space for subsequent processes.
Catalogs, as the skeleton of documents, can naturally locate coarse information by searching the leading section titles.
As exemplified in Figure~\ref{fig:ced_example}, the debt balance ``474.860 billion yuan'' appears in only one segment in the credit rating report that is 30 to 40 pages long.
Taking the whole document into Information Extraction (IE) systems is not practical in this condition.
By searching the catalog tree, this entity can be located in the ``Government Debt Situation'' section with prior knowledge.
Unfortunately, most documents are in plain text and do not contain catalogs in an easily accessible format.
Thus, we propose the Catalog Extraction from Documents (CED) task as a preliminary step to any extremely long document-level IE tasks.
In this manner, fine-grained entities, relations, and events can be further extracted within paragraphs instead of the entire document, which is pragmatic in document-level entity relationship extraction \cite{peng-etal-2017-cross,yu-etal-2020-dialogue,yao-etal-2019-docred} and document-level event extraction \cite{chen-etal-2017-automatically}.

Designing handcrafted heuristics may be a partial solution to the automatic catalog extraction problem.
However, the performance is limited due to three major challenges: 
1) Section titles vary across documents, and there are almost no common rules.
For documents that are in the same format or inherited from the same template, the patterns of section titles are relatively fixed.
Therefore, it is common to use regular expression matching to obtain the whole catalog.
However, such handcrafted heuristics are not reusable when the formats of documents change, and researchers have to design new patterns from scratch, making catalog extraction laborious.
2) Catalogs have deep hierarchies with five- to six-level section headings.
As the level of section headings deepens, titles become increasingly complex, and simple rule systems usually cannot handle fine-grained deep section headings well.
3) A complete sentence may be cut into multiple segments due to mistakes in data acquisition tools.
For example, Optical Character Recognition (OCR) systems are commonly used for obtaining document texts.
However, these systems often make mistakes, and sentences may be incorrectly cut into several segments by line breaks.
These challenges increase the difficulties of using handcrafted rules.

To address the CED task, we first construct a corpus with a total of 650 manually annotated documents.
The corpus includes bid announcements, financial announcements, and credit rating reports.
These three types of documents vary in length and catalog complexity.
This corpus is able to serve as a benchmark for the evaluation of CED systems.
Among these three sources, bid announcements are the shortest in length with simple catalog structures, and financial announcements contain multifarious heading formats, while credit rating reports have deep and nested catalog structures.
In addition, we collect documents from Wikipedia with catalog structures as a large-scale corpus for general model pre-training to enhance the transfer learning ability.
These four types of data cover the first two challenges in catalog extraction.
We also chunk sentences to simulate the incorrect segmentation problem observed in OCR systems, which covers the third challenge in CED.

Based on the constructed dataset, we design a transition-based framework for the CED task.
The catalog tree is formulated as a stack and texts are encased in an input queue.
These two buffers are used to help make action predictions, where each action stands for a control signal that manipulates the composition of a catalog tree.
By constantly comparing the top element of the catalog stack with one text piece from the input queue, the catalog tree is constructed while action predictions are obtained.
The final experimental results show that our method achieves promising results and outperforms other baseline systems.
Besides, the model pre-trained on Wikipedia data is able to transfer the learned information to other domains when training data are limited.

Our contributions are summarized as follows:
\begin{itemize}
    \item We propose a new task to extract catalogs from long documents.
    \item We build a manually annotated corpus for the CED task, together with a large-scale Wikipedia corpus with catalog structures for pre-training. The experimental results show the efficacy of low-resource transfer.
    \item We design a transition-based framework for the task. To the best of our knowledge, this is the first system that extracts catalogs from plain text segments without handcrafted patterns.
\end{itemize}

\section{Related Work}

Since CED is a new task that has not been widely studied, in this section, we mainly introduce approaches applied to similar tasks below.

\textbf{Parsing Problems:}
Similar to other text-to-structure tasks, CED can be recognized as a parsing problem.
A common practice to build syntactic parsers is biaffine-based frameworks with delicate decoding algorithms (e.g., CKY, Eisner, MST) to obtain global optima \cite{biaffine,zhang-etal-2020-treecrf}.
However, when the problem shifts from sentences to documents, former token-wise encoding and decoding methods become less applicable.
As to documents, there are also many popular discourse parsing theories \cite{kamp2013discourse,rhetorical-theory,Theme-rheme}, which aim to extract the inner semantics among Elementary Discourse Units (EDU).
However, the number of EDUs in current corpora is small.
For instance, in the popular RST-DT corpus, the average number of EDU is only 55.6 per document \cite{zhang_top-down_2020}.
When the number of EDUs grows larger, the transition-based method becomes a popular choice \cite{ji-eisenstein-2014-representation}.
Our proposed CED task is based on naive catalog structures that are similar to syntactic structures, but some traditional parsing mechanisms are not suitable since one document may contain thousands of segments.
To this end, we utilize the transition-based method to deal with the CED task.

\textbf{Transition-based Applications:}
The transition-based method parses texts to structured trees in a bottom-up style, which is fast and applicable for extremely long documents.
Despite the successful applications in syntactic and discourse parsing \cite{zhu2013fast,ji-eisenstein-2014-representation,dyer2015transition}, transition-based methods are widely used in information extraction tasks with particular actions, such as Chinese word segmentation \cite{zhang2016transition}, discontinuous named entity recognition \cite{dai-etal-2020-effective} and event extraction \cite{transition-based-ner-events}.
Considering all the characteristics of the CED task, we propose a transition-based method to parse documents into catalog trees.

\section{Dataset Construction}

In this section, we introduce our constructed dataset, the ChCatExt.
Specifically, we first elaborate on the pre-processing, annotation and post-processing methods, then we provide detailed data statistics.

\subsection{Processing \& Annotation}

We collect three types of documents to construct the proposed dataset, including bid announcements\footnote{\url{http://ggzy.hebei.gov.cn/hbjyzx}}, financial announcements\footnote{\url{http://www.cninfo.com.cn}} and credit rating reports\footnote{\url{https://www.chinaratings.com.cn} and \url{https://www.dfratings.com}}.
We adopt Acrobat PDF reader to convert PDF files into docx format and use Office Word to make annotations.
Annotators are required to: 
1) remove running titles, footers (e.g., page numbers), tables and figures;
2) annotate all headings with different outline styles;
and 3) merge mis-segmented paragraphs.
To reduce the annotation bias, each document is assigned to two annotators, and an expert will check the annotations and make final decisions in case of any disagreement.
Due to the length and structure variations, one document may take up to twenty minutes for an annotator to label.

After the annotation process, we use pandoc\footnote{\url{https://pandoc.org}} and additional scripts to parse these files into program-friendly JSON objects.
We insert a pseudo root node before each object to ensure that every document object has only one root.
In real applications, documents are usually in PDF formats, which are immutable and often image-based.
Using OCR tools to extract text contents from those files is a common practice.
However, the OCR tools often split a natural sentence apart when a sentence is physically cut by line breaks or page turnings in PDF, as shown in Figure~\ref{fig:ced_example}.
To simulate real-world scenarios, we randomly sample some paragraphs with a probability of 50\% and chunk them into segments.
For heading strings, we chunk them into segments with lengths of 7 to 20 with jieba\footnote{\url{https://github.com/fxsjy/jieba}} assistance.
This makes heading segmenting more natural, for example, \begin{CJK}{UTF8}{gbsn} ``招标公告'' \end{CJK} will be split into \begin{CJK}{UTF8}{gbsn} ``招标 (zhao biao)'' and ``公告 (gong gao)'' \end{CJK} instead of \begin{CJK}{UTF8}{gbsn} ``招 (zhao)'' and ``标公告 (biao gong gao)'' \end{CJK}.
For other normal texts, we split them into random target lengths between 70 and 100.
Since the workflow is rather complicated, we will open-source all the processing scripts to help further development.

In addition to the above manually annotated data, we collect 665,355 documents from Wikipedia\footnote{\url{https://dumps.wikimedia.org/zhwiki/20211220/}} for model pre-training.
Most of these documents are shallow in catalog structures and short in text lengths.
We keep documents with a catalog depth from 2 to 4 to reach higher data complexity, so that these documents are more similar to the manually annotated ones.
After that, 214,989 documents are obtained.
We chunk these documents in the same way as the manually annotated ones to simulate OCR segmentation.

\subsection{Data Statistics}\label{sec:data_statistics}

\begin{table}[t]
    \centering
    \caption{Data statistics. BidAnn refers to bid announcements, FinAnn is financial announcements and CreRat is credit rating reports. One node may contain multiple segments in its content, and we list the number of nodes here. Depth represents the depth of the document catalog tree (text nodes are also included). Length is obtained by counting the number of document characters.}
    \begin{tabular}{l|r|r|r|r|r|r}
        \hline
        \multirow{2}{*}{Source} & \multirow{2}{*}{\#Docs} & \multirow{2}{*}{Avg.Length} & \multicolumn{3}{|c|}{Avg.\#Nodes} & \multirow{2}{*}{Avg.Depth} \\
        \cline{4-6}
         & & & Heading & Text & Total & \\
         \hline
         BidAnn & 100 & 1,756.76 & 8.04 & 30.61 & 38.65 & 3.00 \\
         FinAnn & 300 & 3,504.22 & 12.09 & 52.31 & 64.40 & 3.79 \\
         CreRat & 250 & 15,003.81 & 27.70 & 81.07 & 108.77 & 4.59 \\
         \hline
         Total ChCatExt & 650 & 7,658.30 & 17.47 & 60.03 & 77.50 & 3.98 \\
         \hline
         Wiki & 214,989 & 1,960.41 & 11.07 & 19.34 & 30.41 & 3.86 \\
         \hline
    \end{tabular}
    \label{tab:data_statistics}
\end{table}

Table~\ref{tab:data_statistics} lists the statistics of the whole dataset.
Among the three types, BidAnn has the shortest length and the shallowest structure, and the headings are similar to each other.
FinAnn is more complex in structure than BidAnn and contains more nodes.
Moreover, there are many forms of headings in FinAnn without obvious patterns, which increases the difficulty of catalog extraction.
CreRat is the most sophisticated one among all types of data.
Its average length is 8.5 times longer than BidAnn while the average depth is 4.59.
However, it contains fewer variations in headings, which may be easier for models to locate.
Compared to manually annotated domain-specific data, Wiki is easy to obtain.
The structure depth is similar to that of FinAnn while its length is 1.5k shorter.
Because of the large size, Wiki is well suited for model pre-training and parameter initializing.

It is worth noting that leaf nodes can be heading or normal texts in catalog trees.
Since normal texts cannot lead a section, all texts are leaf nodes in catalogs.
However, headings could also be leaf nodes if the leading section has no children.
Such a phenomenon appears in approximately 24\% of documents.
Therefore one node cannot be recognized as a text node simply by the number of children, which makes the CED task more complicated.


\section{Transition-based Catalog Extraction}

In this section, we introduce details of our proposed TRAnsition-based Catalog trEe constRuction method \textit{TRACER}.
We first describe the transition-based process, and then introduce the model architecture.

\subsection{Actions \& Transition Process}\label{sec:transition_process}

\begin{table}[t]
    \centering
    \caption{An example of transition-based catalog tree construction. Elements in {\color{red}\textbf{red bold}} represent the current stack top $s$, and elements in {\color{blue}\underline{blue underline}} represent the input text $q$. \$ means the terminal of $\mathcal{Q}$ and the finale of action prediction.}
    \scalebox{0.95}{
    \begin{tabular}{r|m{5.5cm}|r|c}
        \hline
        Step & Catalog Tree Stack $\mathcal{S}$ & Input Queue $\mathcal{Q}$ & Predicted Action \\
        \hline
        \hline
        1 & \redb{Root} & \blueu{Credit Rating Report}, $\ldots$ & Sub-Heading \\
        \hline
        2 & Root [ \redb{Credit Rating Report} ] & \blueu{Debt Situation}, $\ldots$ & Sub-Heading \\
        \hline
        3 & Root [ Credit Rating Report [ \redb{Debt Situation} ] ] & \blueu{The balance}, $\ldots$ & Sub-Text \\
        \hline
        4 & Root [ Credit Rating Report [ Debt Situation [ \redb{The balance} ] ] ] & \blueu{was 474 billion yuan.}, $\ldots$ & Concat \\
        \hline
        5 & Root [ Credit Rating Report [ Debt Situation [ \redb{The balance was 474 billion yuan.} ] ] ] & \blueu{Security Analysis}, $\ldots$ & Reduce \\
        \hline
        6 & Root [ Credit Rating Report [ \redb{Debt Situation} [ The balance was 474 billion yuan. ] ] ] & \blueu{Security Analysis}, $\ldots$ & Reduce \\
        \hline
        7 & Root [ \redb{Credit Rating Report} [ Debt Situation [ The balance was 474 billion yuan. ] ] ] & \blueu{Security Analysis}, $\ldots$ & Sub-Heading \\
        \hline
        8 & Root [ Credit Rating Report [ Debt Situation [ The balance was 474 billion yuan. ] \redb{Security Analysis} ] ] & \blueu{Texts}, \$ & Sub-Text \\
        \hline
        9 & Root [ Credit Rating Report [ Debt Situation [ The balance was 474 billion yuan. ] Security Analysis [ \redb{Texts} ] ] ] & \$ & \$ \\
        \hline
    \end{tabular}
    }
    \label{tab:transition_example}
\end{table}

The transition-based method is designed for parsing trees from extremely long texts.
Since the average length of our CreRat documents is approximately 15k Chinese characters, popular global optimized tree algorithms are apparently too costly to be utilized here.

Action design plays an important role in our transition-based method.
There are two buffers here: 1) the input queue $\mathcal{Q}$ providing one text segment $q$ at each time; and 2) the tree buffer $\mathcal{S}$ that records the final catalog tree, where the current stack top points to $s$.
Actions are obtained by comparing $s$ and $q$ continuously, which results in the buffer changing.
As the comparison process continues, actions compose a control sequence to build the target catalog tree simultaneously.

To solve the mentioned challenges, actions are designed to distinguish between headings and texts. Our actions can also capture the difference between headings from adjacent depth levels. In this way, we construct the catalog tree without regard to its depth and complexity. Additionally, we propose an additional action for text segment concatenation.
Based on these facts, we design 4 actions as follows:
\begin{itemize}
    \item Sub-Heading: current input text $q$ is a child heading node of $s$;
    \item Sub-Text: current input text $q$ is a child text node of $s$;
    \item Concat: current input text $q$ is the latter part of $s$ and their contents should be concatenated;
    \item Reduce: the level of $q$ is above or at the same level as $s$, and $s$ should be updated to its parent node.
\end{itemize}

An example is provided in Table~\ref{tab:transition_example}.
To start the prediction, a \textit{Root} node is given in advance.
The first heading \textit{Credit Rating Report} is regarded as a child of \textit{Root}. Then, \textit{Debt Situation} becomes another heading node. After that, the \textbf{Sub-Text} action suggests that \textit{The balance} is the child node of \textit{Debt Situation} as the body text. Action \textbf{Concat} concatenates two body text. Next, action \textbf{Reduce} leads to the second layer from the third one. We can eventually build a catalog tree with such a sequence of actions.
Furthermore, we present two constraints to avoid illegal results.
The first one is that the action between \textit{Root} node and the first input $q$ can only be Sub-Heading or Sub-Text; Another constraint restricts text nodes to be leaf nodes in the tree, and only Reduce and Concat actions are allowed when $s$ is not a heading.
If the predicted action is illegal, we take the second-best prediction as the final result.

\subsection{Model Architecture}

\begin{figure}[t]
    \centering
    \includegraphics[width=\textwidth]{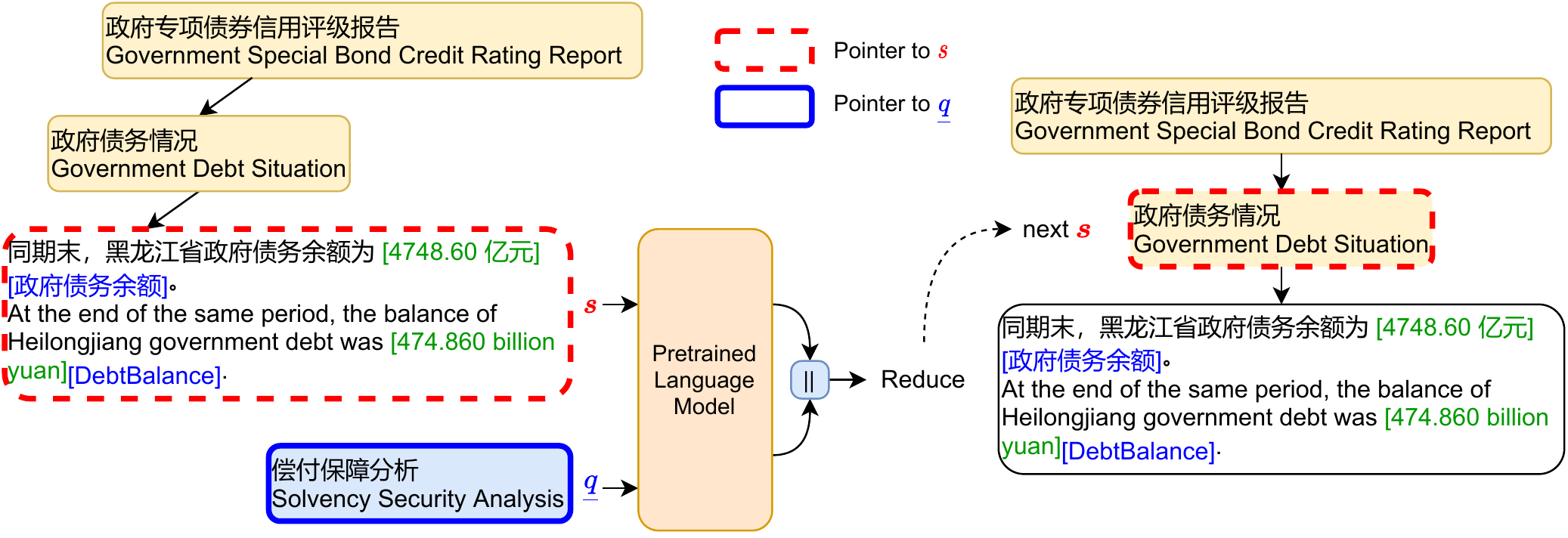}
    \caption{Framework of the transition-based catalog extraction.}
    \label{fig:tracer_framework}
\end{figure}

As Figure~\ref{fig:tracer_framework} shows, the given inputs $s$ and $q$ are encoded via a pre-trained language model (PLM).
Here, we use a light version of Chinese whole word masking RoBERTa (RBT3) \cite{cui-bert-wwm} to obtain encoded representations $\boldsymbol{s}$ and $\boldsymbol{q}$.
After concatenation, $\boldsymbol{g} = \boldsymbol{s} || \boldsymbol{q}$ is fed into Feed-Forward Networks (FFN).
The FFN is composed of two linear transform layers with ReLU activation function and dropout.
Finally we adopt the softmax function to obtain the predicted probabilities as shown in Equation~\ref{eqn:prob}.
\begin{align}
\label{eqn:prob}
    \boldsymbol{o} &= \textrm{FFN}(\boldsymbol{g}), \nonumber\\
    p(\mathcal{A}|s,q) &= \textrm{softmax}(\boldsymbol{o}),
\end{align}
where $\mathcal{A}$ denotes all the action candidates. In this way, we can capture the implicit semantic relationship between two nodes.

During prediction, we take the action with maximal probability $p$ as the predicted result:
\begin{equation}
    \nonumber
    a_i = \argmax_{a\in\mathcal{A}} p(\mathcal{A}|s,q),
\end{equation}
where $a_i \in \mathcal{A}$ is the predicted action.
As discussed in \S~\ref{sec:transition_process}, we use two extra constraints to help force decoding legal action results.
If $a_i$ is an illegal action, we sort the predicted probabilities in reverse order, and then find the legal result with the highest probability.

As for training, we take cross entropy as the loss function to help update the model parameters:
\begin{equation}
    \nonumber
    \mathcal{L} = -\sum_i \mathbb{I}_{y_a = a_i} \log p(a_i|s, q),
\end{equation}
where $\mathbb{I}$ is the indicator function, $y_a$ is the gold action, and $a_i \in \mathcal{A}$ is the predicted action.

\section{Experiments}
\subsection{Datasets}
We further split the datasets into train, development, and test sets with a proportion of 8:1:1 for training.
To fully utilize the scale advantage of the Wiki corpus, we use it to train the model for 40k steps and subsequently save the PLM parameters for transferring experiments.

\subsection{Evaluation Metrics}\label{sec:exp_metrics}

We use the overall micro F1 score on predicted tree nodes to evaluate performances.
Each node in a tree can be formulated as a tuple: (level, type, content), where level refers to the depth of the current node, type refers to the node type (either \textit{Heading} or \textit{Text}), and content refers to the string that the node carries.
The F1 score can be obtained by comparing gold tuples and predicted tuples.

\begin{equation}
    \nonumber
    P = \frac{N_{r}}{N_{p}},\quad R = \frac{N_{r}}{N_{g}},\quad F1 = \frac{2PR}{P+R},
\end{equation}
where $N_{r}$ denotes the number of correctly matched tuples, $N_{g}$ represents the number of gold tuples and $N_{p}$ denotes the number of predicted tuples.

\subsection{Baselines}

Few studies focus on the catalog extraction task, thus we propose two baselines for objective comparisons.

\textbf{1) Classification Pipeline:} 
The catalog extraction task can be formulated in two steps: segment concatenation and tree prediction.
For the first step, we take the text pairs as input and adopt the [CLS] token representation to predict the concatenation results.
Suppose the depth of a tree is limited, the depth level can be regarded as a classification task with $\textrm{MaxHeadingDepth} + 1$ labels, where ``1'' stands for the text node label.
We use PLM with TextCNN \cite{textcnn} to make level predictions.

\textbf{2) Tagging:}
Inheriting the idea of two-step classification from above, the whole task can be formulated as a tagging task.
The segment concatenation sub-task reflects the BIO tagging scheme, and the level depth and node type are tagging labels.
We use PLM with LSTM and CRF to address this tagging task.

\subsection{Experiment Settings}

Experiments are conducted with an NVIDIA TitanXp GPU.
We use RBT3\footnote{\url{https://huggingface.co/hfl/rbt3}}, a Chinese RoBERTa variation, as the PLM.
We use AdamW \cite{adamw} to optimize the model with a learning rate of 2e-5.
Models are trained for 10 epochs.
The training batch size is 20, and the dropout rate is 0.5.
We take 5 trials with different random seeds for each experiment and report average results on the test set with the best model evaluated on the development set.
For the classification pipeline and the tagging baselines, we set the maximal heading depth to 8.

\subsection{Main Results}

\begin{table}[t]
    \centering
    \caption{Main results on ChCatExt. The scores are calculated via the method described in \S~\ref{sec:exp_metrics}. Please beware the overall scores are NOT the average of Heading and Text scores. Heading and Text scores are obtained from a subset of predicted tuple results, where all the node types are ``Heading'' or ``Text''. The overall scores are calculated from the universal set, so they are often lower than the Heading and Text scores. WikiBert represents the PLM that is trained on the wiki corpus in advance.}
    \scalebox{0.9}{
    \begin{tabular}{l|ccc|ccc|ccc}
        \hline
        \multirow{2}{*}{Methods} & \multicolumn{3}{|c|}{Heading} & \multicolumn{3}{|c|}{Text} & \multicolumn{3}{|c}{Overall} \\
        & P & R & F1 & P & R & F1 & P & R & F1 \\
        \hline
        Pipeline & 88.637 & 86.595 & 87.601 & 81.627 & 82.475 & 82.047 & 76.837 & 77.338 & 77.085 \\
        Tagging & 87.456 & 88.241 & 87.846 & 81.079 & 81.611 & 81.344 & 77.746 & 78.800 & 78.269 \\
        \hline
        TRACER & \textbf{90.634} & \textbf{90.341} & \textbf{90.486} & 83.031 & \textbf{85.673} & \textbf{84.328} & \textbf{81.017} & \textbf{83.818} & \textbf{82.390} \\
        \quad w/o Constraints & 89.911 & 89.713 & 89.811 & 82.491 & 84.948 & 83.698 & 80.216 & 83.035 & 81.596 \\
        \hline
        TRACER w/ WikiBert & 88.671 & 89.785 & 89.221 & \textbf{83.308} & 85.025 & 84.156 & 80.820 & 83.357 & 82.063 \\
        \hline
    \end{tabular}
    }
    \label{tab:main_results}
\end{table}

From Table~\ref{tab:main_results}, we find that our proposed TRACER outperforms the classification pipeline and tagging baselines by 5.305\% and 4.121\% overall F1 scores.
The pipeline method requires two separate steps to reveal catalog trees, which may accumulate errors across modules and lead to an overall performance decline.
Although the tagging method is a stronger baseline than the pipeline one, it still cannot match TRACER.
The reason may be the granularities that these methods focus on.
The pipeline and the tagging methods directly predict the depth level for each node, while TRACER pays attention to the structural relationships between each node pair.
Besides, since the two baselines need a set of predefined node depth labels, TRACER is more flexible and can predict deeper and more complex structures.

As discussed in \S~\ref{sec:transition_process}, we use two additional constraints to prevent TRACER from generating illegal trees.
The significance of these constraints is presented in the last line of Table~\ref{tab:main_results}.
If we remove them, the overall F1 score drops 0.794\%.
The decline is expected, but the variation is small, which shows the robustness of the TRACER model design.

Interestingly, the PLM trained on the Wiki corpus does not bring performance improvements as expected.
This may be due to the different data distributions between Wikipedia and our manually annotated ChCatExt.
The following transferring analysis section \S~\ref{sec:transfer_exp} contains more results with WikiBert.

\subsection{Analysis of Transfer Ability}\label{sec:transfer_exp}


\begin{table}[t]
    \centering
    \caption{Transferring F1 results: train on the source set, evaluate on the target set. Training on Wiki is the process of obtaining the WikiBert, so results in TRACER w/ WikiBert are absent since the experiments are duplicates.}
    \begin{tabularx}{1.0\textwidth}{l|RRRR||RRRR}
        \hline
        \multirow{2}{*}{tgt ↓ src →} & \multicolumn{4}{c||}{TRACER} & \multicolumn{4}{c}{TRACER w/ WikiBert} \\
        & BidAnn & FinAnn & CreRat & Wiki & BidAnn & FinAnn & CreRat & Wiki \\
        \hline
        BidAnn & 88.076 & \textbf{25.557} & 8.400 & 2.703 & \textbf{88.200} & 25.260 & \textbf{11.741} & -  \\ \hline
        FinAnn & 7.391 & \textbf{69.249} & 15.543 & 11.388 & \textbf{8.100} & 68.588 & \textbf{20.174} & -  \\ \hline
        CreRat & 2.361 & 14.420 & \textbf{92.790} & 14.029 & \textbf{7.000} & \textbf{30.821} & 92.290 & -  \\ \hline
        \hline
    \end{tabularx}
    \label{tab:transfer_results}
\end{table}

\begin{table}[t]
\centering
\caption{Transfer F1 results: train on k documents from the source set, evaluate on the whole target set.}
\label{tab:transfer_tkset}
\begin{tabularx}{\textwidth}{l|RRR|RRR|RRR}
\hline
\multirow{3}{*}{tgt ↓ src →} & \multicolumn{9}{c}{TRACER}\\
\cline{2-10}      
& \multicolumn{3}{c|}{BidAnn} & \multicolumn{3}{c|}{FinAnn} & \multicolumn{3}{c}{CreRat} \\
& k=3 & 5 & 10 & k=3 & 5 & 10 & k=3 & 5 & 10 \\
\hline
BidAnn & 63.033 & 10.969 & 7.242 & 0.713 & 20.798 & 9.164 & 0.000 & 11.490 & \textbf{39.264}  \\
FinAnn & 63.460 & \textbf{17.758} & 10.613 & 0.815 & 28.177 & 11.755 & 0.047 & 11.337 & 48.543 \\
CreRat & 77.259 & 14.363 & 14.845 & 1.725 & 25.110 & \textbf{12.636} & 3.255 & 10.768 & 70.277 \\
\hline
\multicolumn{10}{c}{TRACER w/ WikiBert} \\
\hline
BidAnn & \textbf{66.644} & \textbf{14.578} & \textbf{18.719} & \textbf{2.355} & \textbf{21.482} & \textbf{18.385} & \textbf{1.024} & \textbf{12.781} & 30.626 \\
FinAnn & \textbf{67.040} & 15.509 & \textbf{15.467} & \textbf{3.515} & \textbf{32.568} & \textbf{14.125} & \textbf{1.765} & \textbf{25.285} & \textbf{56.192} \\
CreRat & \textbf{79.029} & \textbf{16.424} & \textbf{14.936} & \textbf{4.517} & \textbf{27.528} & 12.398 & \textbf{18.659} & \textbf{19.238} & 67.775 \\
\hline
\end{tabularx}
\end{table}


\begin{table}[t]
\centering
\caption{Transfer F1 results: train on the source set, further train on k target documents, evaluate on the target set.}
\label{tab:transfer_tstket}
\begin{tabularx}{\textwidth}{l|RRR|RRR|RRR}
\hline
\multirow{3}{*}{src ↓ tgt →} & \multicolumn{9}{c}{TRACER}\\
\cline{2-10}      
& \multicolumn{3}{c|}{BidAnn} & \multicolumn{3}{c|}{FinAnn} & \multicolumn{3}{c}{CreRat} \\
& k=3 & 5 & 10 & k=3 & 5 & 10 & k=3 & 5 & 10 \\
\hline
BidAnn & - & 87.995 & 74.630 & \textbf{26.640} & - & 29.607 & \textbf{37.991} & \textbf{56.658} & - \\
FinAnn & - & 87.991 & 75.921 & 24.502 & - & \textbf{35.672} & \textbf{38.560} & \textbf{68.287} & - \\
CreRat & - & \textbf{88.923} & 79.061 & 27.988 & - & 43.066 & 52.139 & 72.954 & - \\
\hline
\multicolumn{10}{c}{TRACER w/ WikiBert} \\
\hline
BidAnn & - & \textbf{91.400} & \textbf{76.626} & 25.709 & - & \textbf{29.729} & 29.762 & 53.406 & - \\
FinAnn & - & \textbf{93.608} & \textbf{76.106} & 28.035 & - & 33.698 & 36.217 & 65.825 & - \\
CreRat & - & 88.777 & \textbf{81.020} &\textbf{ 32.345} & - & \textbf{45.488} & \textbf{57.580} & \textbf{73.519} & - \\
\hline
\end{tabularx}
\end{table}


\begin{table}[t]
\centering
\caption{Transfer F1 results: concatenate the source set with k target documents, train on the merged set, evaluate on the target set.}
\label{tab:transfer_tsktet}
\begin{tabularx}{\textwidth}{l|RRR|RRR|RRR}
\hline
\multirow{3}{*}{src ↓ tgt →} & \multicolumn{9}{c}{TRACER}\\
\cline{2-10}      
& \multicolumn{3}{c|}{BidAnn} & \multicolumn{3}{c|}{FinAnn} & \multicolumn{3}{c}{CreRat} \\
& k=3 & 5 & 10 & k=3 & 5 & 10 & k=3 & 5 & 10 \\
\hline
BidAnn & - & \textbf{80.924} & 73.703 & 27.237 & - & 29.528 & \textbf{45.813} & \textbf{56.273} & - \\
FinAnn & - & \textbf{88.902} & 76.137 & 24.800 & - & 31.989 & 32.173 & 22.583 & - \\
CreRat & - & \textbf{88.310} & \textbf{82.551} & \textbf{31.768} & - & \textbf{45.847} & \textbf{61.107} & \textbf{73.933} & - \\
\hline
\multicolumn{10}{c}{TRACER w/ WikiBert} \\
\hline
BidAnn & - & 78.647 & \textbf{79.606} & \textbf{28.243} & - & \textbf{33.735} & 45.226 & 55.887 & - \\
FinAnn & - & 83.227 & \textbf{76.556} & \textbf{30.823} & - & \textbf{34.878} & \textbf{36.559} & \textbf{62.070} & - \\
CreRat & - & 83.823 & 76.442 & 29.587 & - & 35.086 & 59.217 & 71.713 & - \\
\hline
\end{tabularx}
\end{table}

One of our motivations for building a model to solve the CED task is that we want to provide a general model that fits all kinds of documents.
Therefore, we conduct transfer experiments under different settings to find the interplay among different sources with diverse structure complexities.
The results are listed in Table~\ref{tab:transfer_results} to~\ref{tab:transfer_tsktet}.

We first train models on three separate source datasets and make direct predictions on target datasets.
From the left part of Table~\ref{tab:transfer_results}, we can obtain a rough intuition of the data distribution.
The model trained on BidAnn makes poor predictions on FinAnn \& CreRat, and gets only 7.391\% and 2.361\% F1 scores, which also conforms with former discussions in \S~\ref{sec:data_statistics}.
BidAnn is the easiest one among the three sources of datasets, so the generalization ability is less robust.
FinAnn is shallower in structure, but it contains more variations.
The model trained on FinAnn only obtains a 69.249\% F1 score evaluated on FinAnn itself.
However, it gets better results on BidAnn (25.557\%) and CreRat (14.420\%) than the others.
The model trained on CreRat gets 92.790\% on itself.
However, it does not generalize well on the other two sources.
We also provide the zero-shot cross-domain results from Wiki to the other three subsets.
Although the results are poor under the zero-shot setting, the pre-trained WikiBert shows great transfer ability.
Comparing results horizontally in Table~\ref{tab:transfer_results}, we find that the pre-trained WikiBert could provide good generalization and outperforms the vanilla TRACER among 6 out of 9 transferring data pairs.
The other 3 pairs' results are very close and competitive.

To further investigate the generalization ability of pre-training on the Wiki corpus, we take an extreme condition into consideration, where only a few documents are available to train a model.
In this case, as shown in Table~\ref{tab:transfer_tkset}, we train models with only k source documents and calculate the final evaluation results on the whole target test set.
Each model is evaluated on the original source development set to select the best model and then the best model makes final predictions on the target test set.
TRACER w/ WikiBert outperforms vanilla TRACER among 23 out of 27 transferring pairs.
There is no obvious upward trend when increasing k from 3 to 10, which is unexpected and suggests that the model may suffer from overfitting problems on such extremely small training sets.

In most cases of real-world applications, a few target documents are available.
Supposing we want to transfer models from source sets to target sets with k target documents available, there are two possible methods to utilize such data.
The first one is to train on the source set, and then further train with k target documents; the other one is to concatenate the source set and k targets into a new train set.
We conduct experiments under these two settings. The results are presented in Table~\ref{tab:transfer_tstket} and \ref{tab:transfer_tsktet}.
Comparing the vanilla TRACER model results, we find that concatenating has 10 out of 18 pairs that outperform the further training method.
From k=3 to 10, there are 2, 3, and 5 pairs that show better results, indicating that the concatenation method is better as k increases.
WikiBert has different effects under these two settings.
In the further training method, WikiBert is more powerful (11 out of 18 pairs), while it is less useful in the concatenation method (8 out of 18 pairs).

Overall, we find that:
1) WikiBert achieves good performances, especially when the training set is small;
2) If there are k target documents available besides the source set, WikiBert is not a must, and concatenating the source set with k targets to make a new train set may produce better results.


\subsection{Analysis on the Number of Training Data}

\begin{figure}[t]
    \centering
    \includegraphics[width=\textwidth]{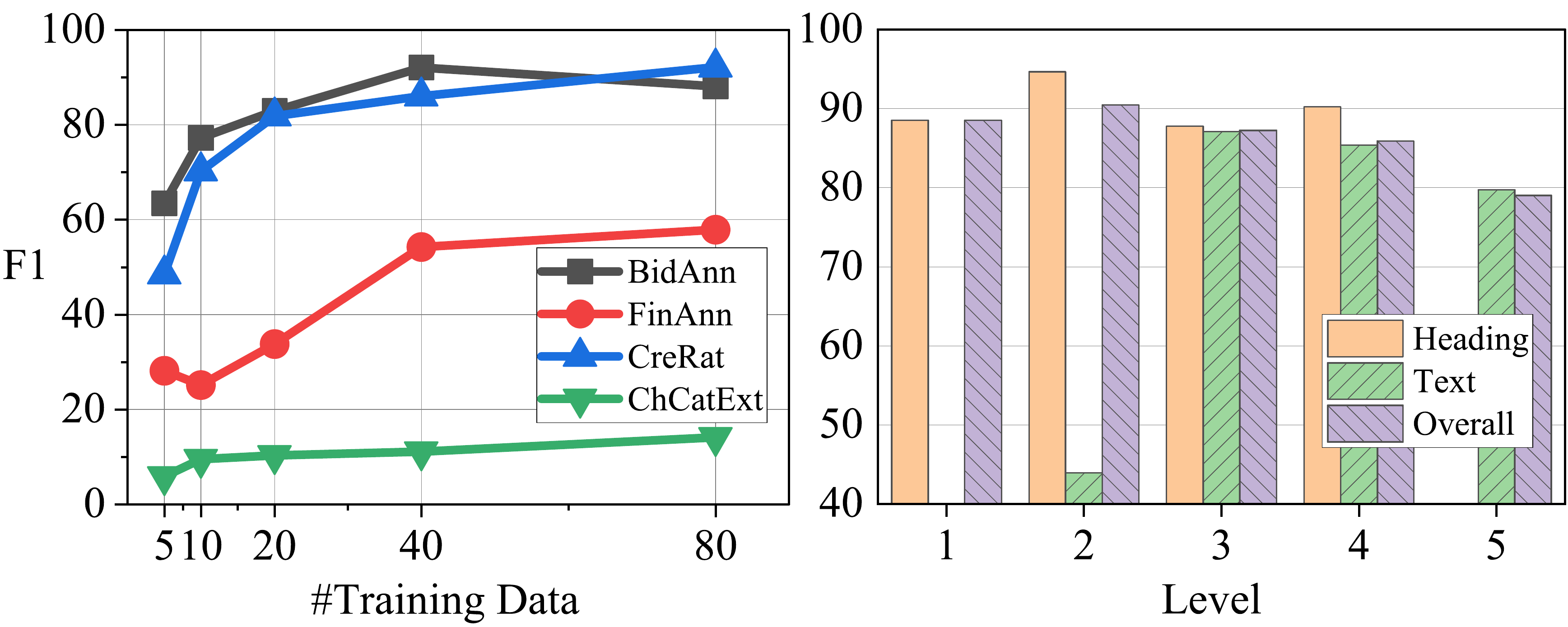}
    \caption{F1 scores with different numbers   training data scales (left) and levels (right). The scale means the number of documents participating in training. The results with different levels are evaluated on ChCatExt.}
    \label{fig:scale_and_level}
\end{figure}

The left part of Figure~\ref{fig:scale_and_level} shows the average results on each separate dataset with different training data scales.
Although BidAnn is the smallest data, the model still gets a 63.460\% F1 score and surpasses the other datasets.
Interestingly, a decline is observed in BidAnn when the number of training documents increases from 40 to 80.
We take it as a normal fluctuation representing a performance saturation since the performance standard deviation is 4.950\% when the training data scale is 40.
Besides, we find that TRACER has good performance on CreRat.
This indicates that TRACER performs well in datasets with deeply nested documents if the catalog heading forms are less varied.
In contrast, TRACER is lower in performance on FinAnn than BidAnn and CreRat, and it is more data-hungry than other data sources.
For ChCatExt, the merged dataset, performance grows slowly with the increase of training data scale, and more data are needed to be fully trained.
Comparing the overall F1 performance of 82.390\% on the whole ChCatExt, the small scale of the training set may lead to a bad generalization.

\subsection{Analysis on Different Depth}

From the right bar plot of Figure~\ref{fig:scale_and_level}, it is interesting to see the F1 scores are 0\% in level 1 text and level 5 heading.
This is mainly due to the golden data distribution characteristics that there are no text nodes in level 1, and there are few headings in deeper levels, leading to zero performances.
The F1 score on level 2 text is only 43.938\%, which is very low compared to the level 3 text result.
Considering that there are only 6.092\% of text nodes among all the level 2 nodes, this indicates that TRACER may be not robust enough.
Combining the above factors, we find that the overall performance increases from level 1 to 2 and then decreases as the level grows deeper.
To reduce the performance decline with deeper levels, additional historical information needs to be considered in future work.

\section{Conclusion and Future Discussion}

In this paper, we build a large dataset for automatic catalog extraction, including three domain-specific subsets with human annotations and large-scale Wikipedia documents with automatically annotated structures.
Based on this dataset, we design a transition-based method to help address the task and get promising results.
We pre-train our model on the Wikipedia documents and conduct experiments to evaluate the transfer learning ability.
We expect that this task and new data could boost the development of Intelligent Document Processing.

We also find some imperfections from the experimental results.
Due to the distribution gaps, pre-training on Wikipedia documents does not bring performance improvements on the domain-specific subsets, although it is proven to be useful under the low-resource transferring settings.
Besides, the current model only compares two single nodes each time and misses the global structural histories.
Better encoding strategies may need to be discovered to help the model deal with deeper structure predictions.
We leave these improvements to future work.

\section*{Acknowledgments}

This work was supported by the National Natural Science Foundation of China (Grant No. 61936010) and Provincial Key Laboratory for Computer Information Processing Technology, Soochow University.
This work was also supported by the Priority Academic Program Development of Jiangsu Higher Education Institutions, and the joint research project of Huawei Cloud and Soochow University.
We would also like to thank the anonymous reviewers for their valuable comments.

%
%
%
\bibliographystyle{splncs04}

\bibliography{references}

\end{document}